\documentclass[conference]{IEEEtran}
\usepackage{cite}
\usepackage{amsmath,amssymb,amsfonts}
\usepackage{algorithm}
\usepackage[noend]{algorithmic}
\usepackage{graphicx}
\usepackage{textcomp}
\usepackage{xcolor}
\usepackage{dsfont}
\def\BibTeX{{\rm B\kern-.05em{\sc i\kern-.025em b}\kern-.08em
 T\kern-.1667em\lower.7ex\hbox{E}\kern-.125emX}}
\usepackage{amsthm}
\theoremstyle{definition}

\begin{document}

\title{Neural Network Topologies for Sparse Training}

\author{\IEEEauthorblockN{
 Ryan A. Robinett$^1$ and Jeremy Kepner$^{1,2}$
}
\vspace{1ex}
\IEEEauthorblockA{
	$^1$MIT Department of Mathematics, $^2$MIT Lincoln Laboratory Supercomputing Center}
}

\maketitle

\begin{abstract}
The sizes of deep neural networks (DNNs) are rapidly outgrowing the capacity of hardware to store and train them. Research over the past few decades has explored the prospect of sparsifying DNNs before, during, and after training by pruning edges from the underlying topology. The resulting neural network is known as a sparse neural network. More recent work has demonstrated the remarkable result that certain sparse DNNs can train to the same precision as dense DNNs at lower runtime and storage cost. An intriguing class of these sparse DNNs is the X-Nets, which are initialized and trained upon a sparse topology with neither reference to a parent dense DNN nor subsequent pruning. We present an algorithm that deterministically generates sparse DNN topologies that, as a whole, are much more diverse than X-Net topologies, while preserving X-Nets' desired characteristics.
\end{abstract}

\begin{IEEEkeywords}
feedforward neural networks, sparse matrices, artificial intelligence
\end{IEEEkeywords}

\section{Introduction}

\let\thefootnote\relax\footnotetext{This material is based in part upon work supported by the NSF under grant number DMS-1312831. Any opinions, findings, and conclusions or recommendations expressed in this material are those of the authors and do not necessarily reflect the views of the National Science Foundation.}

As research in artificial neural networks progresses, the sizes of state-of-the-art deep neural network (DNN) architectures put increasing strain on the hardware needed to implement them \cite{7298594, kepner_exact}. In the interest of reduced storage and runtime costs, much research over the past decade has focused on the sparsification of artificial neural nets \cite{lecun1990optimal,hassibi1993second,srivastava2014dropout,iandola2016squeezenet,DBLP:journals/corr/SrinivasB15,DBLP:journals/corr/HanMD15,7298681,KepnerGilbert2011,kepner2017enabling,kumar2018ibm,kepner2018mathematics}. In the listed resources alone, the methodology of sparsification includes Hessian-based pruning \cite{lecun1990optimal,hassibi1993second}, Hebbian pruning  \cite{srivastava2014dropout}, matrix decomposition in \cite{7298681}, and graph techniques \cite{kumar2018ibm,KepnerGilbert2011,kepner2017enabling,kepner2018mathematics}. Yet all of these implementations are alike in that a DNN is initialized and trained, and then edges deemed unnecessary by certain criteria are pruned.

Unlike most strategies for creating sparse DNNs, the X-Net strategy presented in \cite{DBLP:journals/corr/abs-1711-08757} is sparse ``\textit{de novo}"---that is, X-Nets are neural networks initialized upon sparse topologies. X-Nets are observed to train as well on various data sets as their dense counterparts, while exhibiting reduced memory usage \cite{DBLP:journals/corr/abs-1711-08757,alford}. Further, by offering sparse alternatives to fully-connected and convolutional layers---X-Linear and X-Conv layers, respectively---X-Nets exhibit such performance on not only generalized DNN tasks, but also image recognition tasks canonically reserved for convolutional neural networks \cite{7298681}.

X-Net layers are constructed using properties of expander graphs \cite{8186802}. Due to the tendency of expander graphs to achieve path-connectedness (see Mathematical Preliminaries), this structure is what enables X-Nets to train to diverse models with the same precision as dense DNNs \cite{DBLP:journals/corr/abs-1711-08757}. Random X-Linear layers achieve path-connectedness probabilistically, while explicit X-Linear layers, constructed from Cayley graphs, aim to achieve path-connectedness deterministically \cite{DBLP:journals/corr/abs-1711-08757}. 
As an artifact of their construction from Cayley graphs, explicit X-Linear layers are required have the same number of nodes as adjacent layers. This constrains the kinds of X-Nets which may be constructed deterministically.

\begin{figure}
 \centering
 \includegraphics[trim={2.5cm 0 2.5cm 0}, clip, width=\columnwidth]{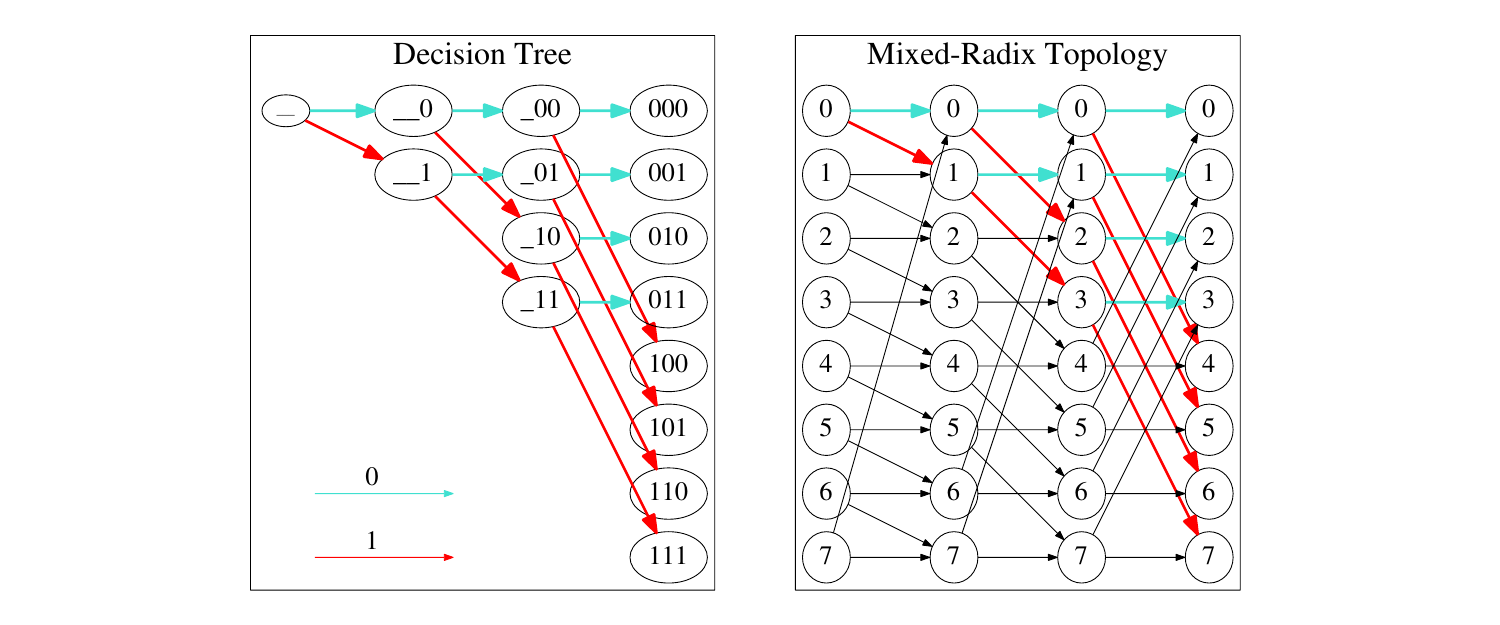}
 \caption{Construction of the mixed-radix topology defined by $\mathcal{N}=(2,2,2)$ using overlapping decision trees. (left) A four-layer binary decision tree. (right) A four-layer mixed radix topology composed of eight offset decision trees.
 }
 \label{fig:Mixed-Radix}
\end{figure}

We propose RadiX-Nets as a new family of \textit{de novo} sparse DNNs that deterministically achieve path-connectedness while allowing for diverse layer architectures. Instead of emulating Cayley graphs, RadiX-Nets achieve sparsity using properties of mixed-radix numeral systems, while allowing for diversity in network topology through the Kronecker product \cite{LOAN200085}. Additionally, RadiX-Nets satisfy symmetry, a property which both guarantees path connectedness and precludes inherent training bias in the underlying sparse DNN architecture.

\begin{figure*}[ht]
 \centering
 \includegraphics[width=7in]{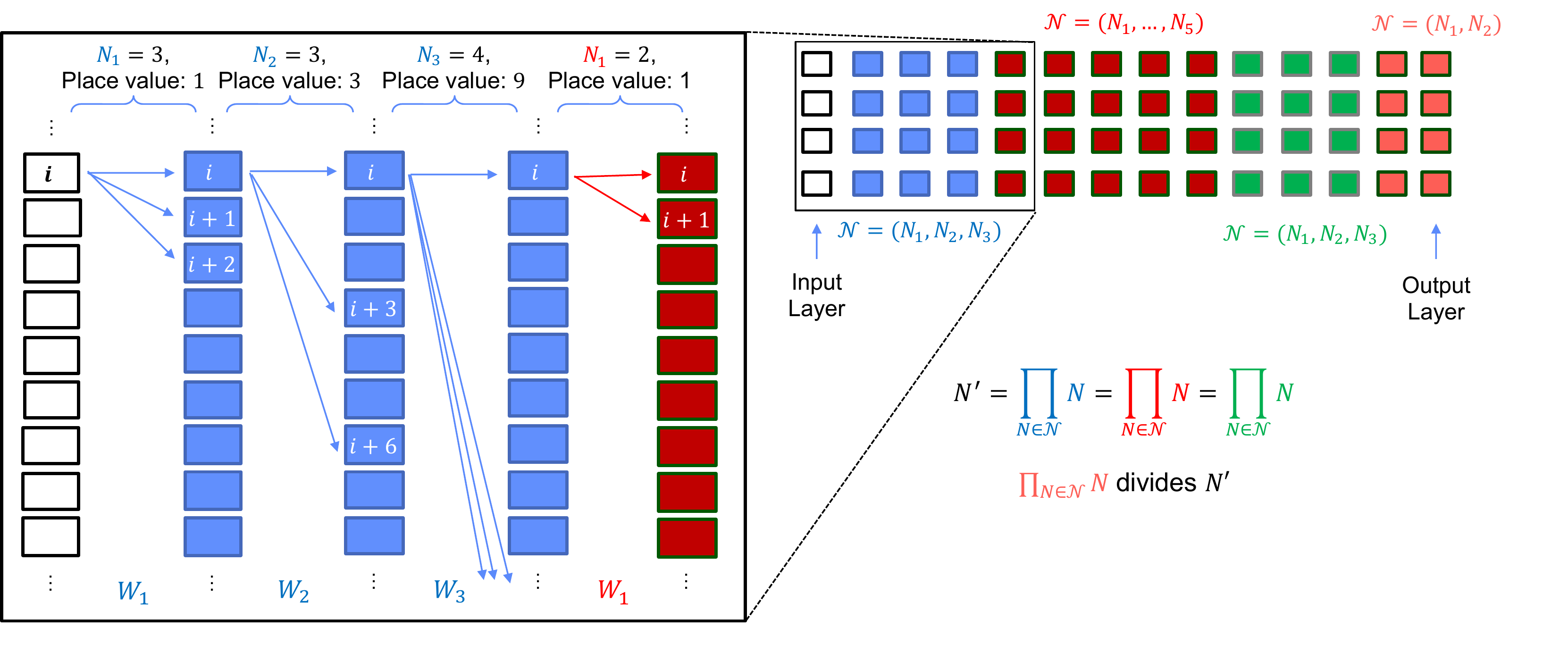}
 \caption{A RadiX-Net \emph{prior} to Kronecker product is a layered graph wherein each layer is a mixed-radix topology. (left) A single mixed-radix topology within a concatenation of mixed-radix topologies, defined by mixed-radix system ${\color{blue}\mathcal{N}=(3,3,4)}$. (top right) A concatenation of the mixed-radix topologies defined by ${\color{blue}\mathcal{N}},{\color{red}\mathcal{N}},{\color{green}\mathcal{N}},$ and ${\color{orange}\mathcal{N}}$. The mixed-radix topologies are concatenated such that the output nodes of one are identified label-wise with the input-nodes of the next. (bottom right) Strict relationships between ${\color{blue}\mathcal{N}},{\color{red}\mathcal{N}},{\color{green}\mathcal{N}},$ and ${\color{orange}\mathcal{N}}$ allow for RadiX-Nets to satisfy sparsity, symmetry, and path-connectedness.}
 \label{fig:Stacking}
\end{figure*}

\section{Mathematical Preliminaries}

\begin{figure*}[ht]
 \centering
 \includegraphics[width=6in]{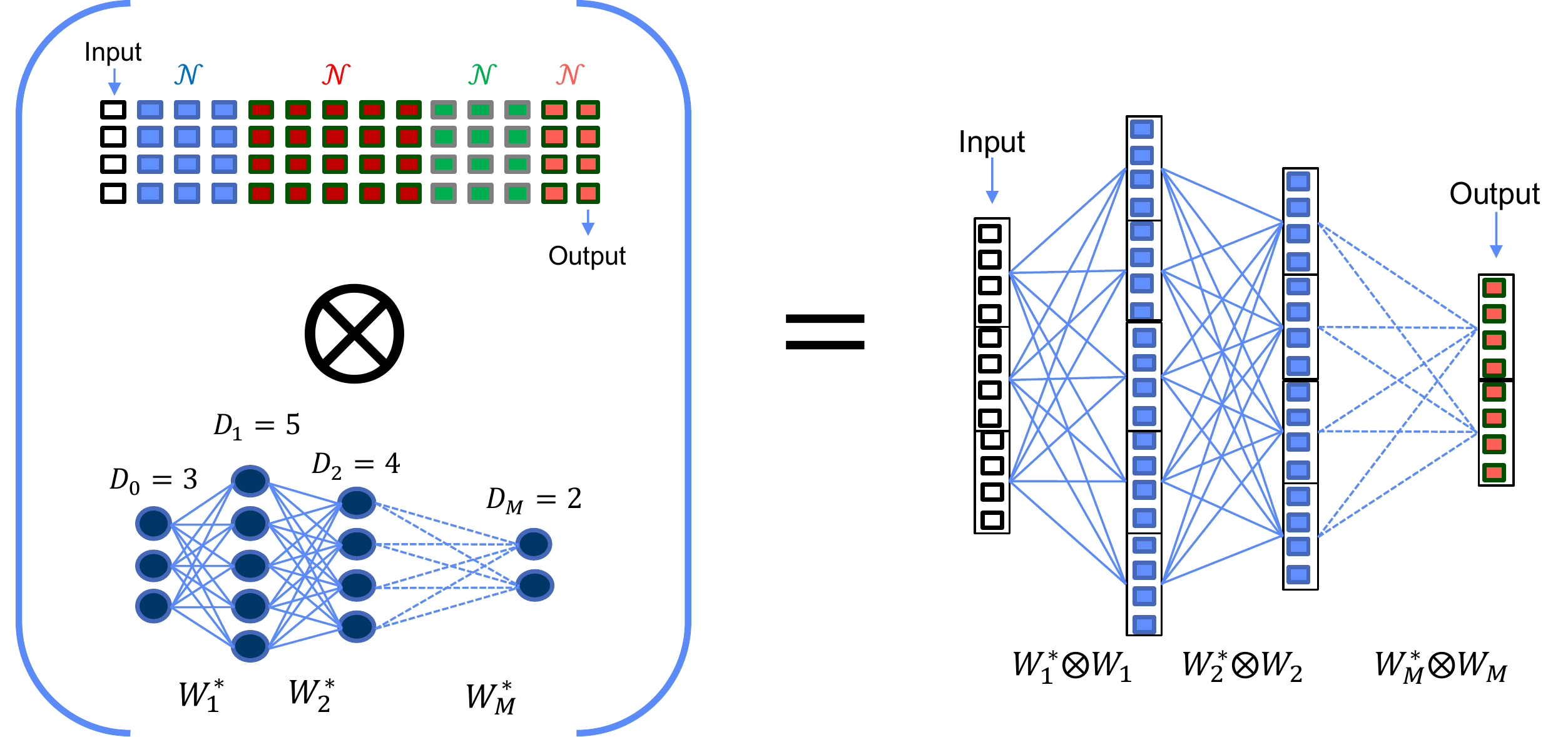}
 \caption{The final step of RadiX-Net construction involves Kronecker products of adjacency submatrices of mixed-radix topologies and adjacency submatrices of an arbitrary dense deep neural network with the same number of layers. The number of vertices in each layer of the dense deep neural networks provides an additional set of parameters by which a wide range of RadiX-Nets can be defined. 
 }
 \label{fig:Kronecker}
\end{figure*}

Understanding RadiX-Nets' graph-theoretic construction and underlying mathematical properties requires defining a few concepts. RadiX-Nets are composed of sub-nets that are herein referred to as mixed-radix topologies. Mixed-radix topologies are based on properties of mixed-radix number systems, and can be constructed from overlapping decision trees (see Figure~\ref{fig:Mixed-Radix}). A mixed-radix numeral system is the sole parameter used to uniquely specify a mixed-radix topology. Mixed-radix topologies are a kind of feedforward neural net topology (FNNT), which is a layered graph wherein all vertices in one layer point only to some number of vertices in the next. The adjacency matrix of an FNNT is uniquely defined by the adjacency submatrices corresponding to each of its layers. Essentially, RadiX-Net topologies are constructed from Kronecker products of mixed-radix adjacency submatrices and dense DNN adjacency submatrices (see Figure~\ref{fig:Kronecker}). The main properties of interest in RadiX-Nets are path-connectedness---which ensures each output depends upon all inputs---and symmetry, which ensures that there is the same number of paths between each input and output.

\emph{Mixed-Radix Numeral System}:
Let $\mathcal{N}=(N_1,\ldots,N_L)$ be an ordered set of $L$ integers greater than 1. Let $N^\prime=\prod_{i=1}^LN_i$. All such $\mathcal{N}$ implicitly define a numeral system which bijectively represents all integers in $\{0,\ldots,N^\prime-1\}$. That is, the set of ordered sets
$$\big\{(n_1,\ldots,n_L)\mid n_i\in\{0,\ldots,N_i-1\}\big\}$$
maps bijectively to $\{0,\ldots,N^\prime-1\}$ by the map
$$(n_1,\ldots,n_L)\longleftrightarrow\sum_{i=1}^L\left(n_i\prod_{j=1}^{i-1}N_j\right).$$
Mixed-radix numeral systems arise naturally in numerous graph-theoretic constructions, such as decision trees (see Figure~\ref{fig:Mixed-Radix}).

\emph{Feedforward Neural Net Topology (FNNT)}: 
An FNNT $G$ with $n+1$ layers of nodes---including input and output layers---is an $(n+1)$-partite directed graph with independent components $U_0,\ldots,U_n$ satisfying the constraints that
\begin{itemize}
\item if there exists an edge from $u\in U_i$ to $v\in U_j$, then $j=i+1$, and
\item the out-degree of $u\in U_i$ is nonzero for all $i<n$.
\end{itemize}

\emph{Adjacency Submatrix of an FNNT}:
Say $G$ is an FNNT. Let $G_i$ be the restriction of $G$ to the set of nodes $U_{i-1}\cup U_i$ and the set of edges from $U_{i-1}$ to $U_i$ in $G$. We define $m_i=\lvert U_{i-1}\rvert$ and $n_i=\lvert U_i\rvert$ for all $i$. Up to a permutation of indices, the adjacency matrix of $G_i$ is of the form
$$\left(\begin{array}{c | c}
\mathbf{0}_{m_i,m_i} & \mathbf{W}_i \\
\hline
\mathbf{0}_{n_i,m_i} & \mathbf{0}_{n_i,n_i} \\
\end{array}\right)$$
for some $\mathbf{W}_i$, where $\mathbf{0}_{a,b}$ is the $a\times b$ matrix of zeros. We refer to $\mathbf{W}_i$ as the adjacency submatrix of the restriction $G_i$.

Conversely, say that an ordered set $\mathcal{W}=(\mathbf{W}_1,\ldots,\mathbf{W}_n)$ of matrices is such that
\begin{itemize}
\item the only nonzero entries of $\mathbf{W}_i$ are ones for all $i$, and
\item no column of $\mathbf{W}_i$ is the zero vector.
\end{itemize}
If the number of columns in $\mathbf{W}_{i-1}$ equals the number of rows in $\mathbf{W}_i$ for all $i\in\{1,\ldots,n\}$, then $\mathcal{W}$ defines a unique FNNT with $n+1$ layers of nodes.

\emph{Path-Connectedness}:
We defined path-connectedness as follows: let $G$ be an FNNT with $n+1$ layers of nodes. $G$ is path-connected if, for every $u\in U_0$ and every $v\in U_n$, there exists a path from $u$ to $v$.

\emph{Symmetry}:
We define symmetry as follows: let $G$ be an FNNT with $n+1$ layers of nodes. $G$ is symmetric if there exists a positive integer $m$ such that, for all $u\in U_0$ and all $v\in U_n$, there exist exactly $m$ paths from $u$ to $v$. If $G$ is symmetric, it is path-connected.

\section{RadiX-Net Topologies}

\begin{figure}
 \centering
 \includegraphics[trim={0 1.8cm 0 1.5cm}, clip, width=8cm]{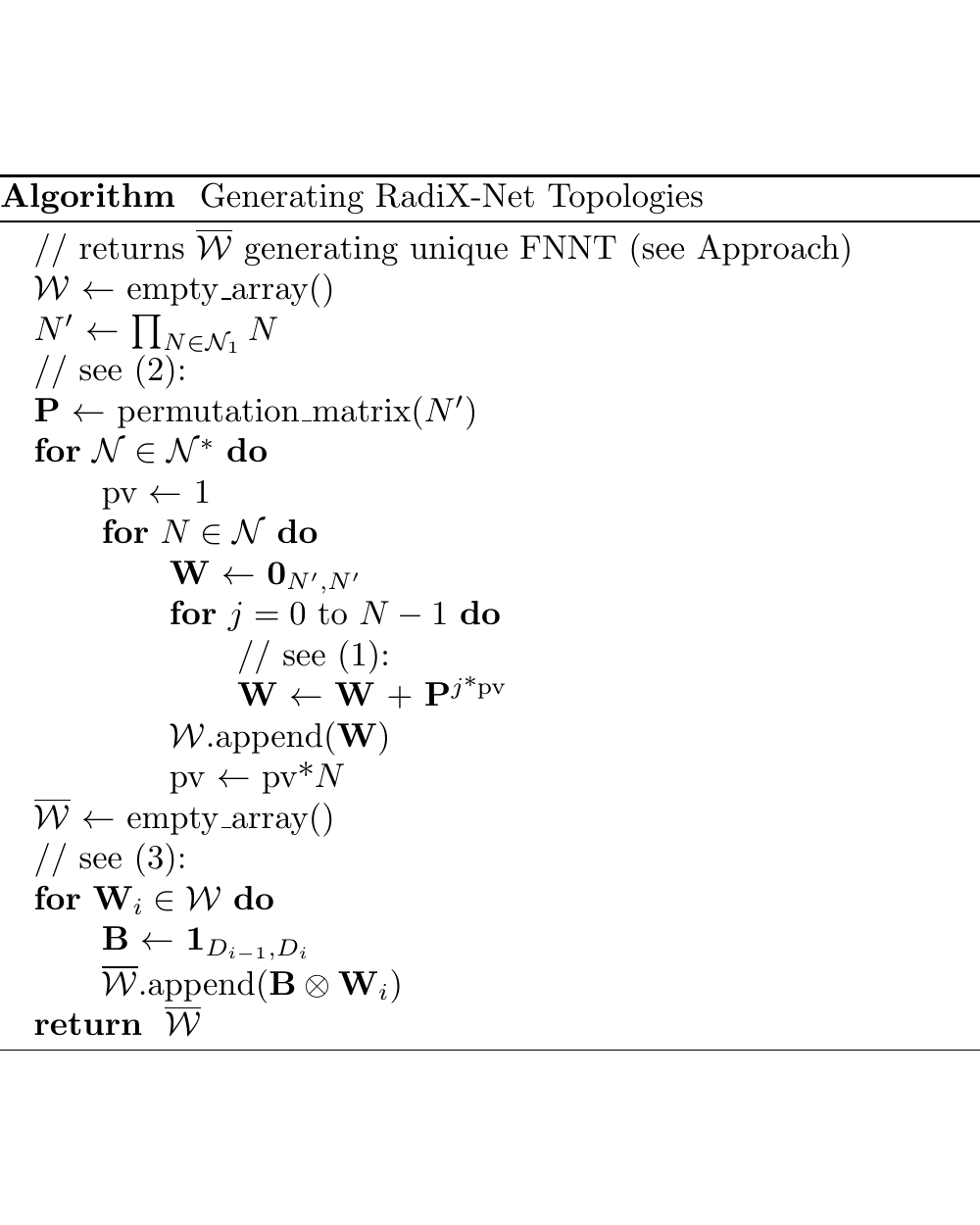}
 \caption{An algorithm for generating the RadiX-Net topology defined by list $\mathcal{N}^*=(\mathcal{N}_1,\ldots,\mathcal{N}_M)$ of mixed-radix numeral systems and list $\mathcal{D}=(D_0,\ldots,D_{\overline{M}})$ of positive integers.}
 \label{fig:pseudo}
\end{figure}

We construct RadiX-Net topologies using mixed-radix topologies as building blocks, as motivated by Figure \ref{fig:Stacking}.

\emph{Mixed-Radix Topologies:}
Let $L$ be a positive integer, and let $\mathcal{N}=(N_1,\ldots,N_L)$, where $N_i$ is an integer greater than one for all $i$. Let $N^\prime=\prod_{N\in\mathcal{N}}N$, and let $U_i$ be a set of $N^\prime$ nodes---with labels $0,\ldots,N^\prime-1$---for all $i\in\{0,\ldots,L\}$. For all $i\in\{1,\ldots,L\}$, we create edges from node $j$ in $U_{i-1}$ to node $j+n\prod_{j=1}^{i-1}N_j\textmd{ (mod }N^\prime)$ in $U_i$ for all $n\in\{0,\ldots,N_i-1\}$. Let $\mathbf{W}_i$ be the adjacency submatrix defining the edges from $U_{i-1}$ to $U_i$. By construction, we have that
\begin{equation} \label{circshift}
\mathbf{W}_i = \sum_{j=0}^{N_i-1}\mathbf{P}^{j\nu_i},
\end{equation}
where $\nu_i=\prod_{k=1}^{i-1}N_k$ and $\mathbf{P}$ is the permutation matrix
\begin{equation} \label{perm}
\left(\begin{array}{ccc|c}
0 & \ldots & 0 & 1 \\
\hline
\ & \ & \ & 0 \\
\ & \mathbf{I}_{N^\prime-1} & \ & \vdots \\
\ & \ & \ & 0
\end{array}\right),
\end{equation}
$\mathbf{I}_n$ being the $n\times n$ identity matrix. We refer to the resulting graph as the mixed-radix topology induced by $\mathcal{N}$.

\emph{Constructing RadiX-Net Topologies:}
Here, we formally construct RadiX-Net topologies using mixed-radix topologies, adjacency submatrices, and the Kronecker product, as motivated by Figure \ref{fig:Kronecker}. For an informal programmatic construction, see Figure \ref{fig:pseudo}.

RadiX-Net topologies are uniquely defined by an ordered set $\mathcal{N}^*=(\mathcal{N}_1,\ldots,\mathcal{N}_M)$ of mixed-radix numeral systems $\mathcal{N}_i=(N_1^i,\ldots,N_{L_i}^i)$ together with an ordered set $\mathcal{D}$ of positive integers. We require that
\begin{itemize}
 \item there exists a positive integer $N^\prime$ such that $N^\prime=\prod_{N\in\mathcal{N}_i}N$ for all $i\in\{1,\ldots,M-1\}$, and
 \item $\prod_{N\in\mathcal{N}_M}N$ divides $N^\prime$.
\end{itemize}
Let $\overline{M}=\sum_{i=1}^{M}L_i$, the total number of radices in $\mathcal{N}^*$; we further require that $\mathcal{D}=(D_0,\ldots,D_{\overline{M}})$ consist of $\overline{M}+1$ integers satisfying $D_i\ll N^\prime$ for all $i$.

We construct a RadiX-Net $G$ using $\mathcal{N}^*$ and $\mathcal{D}$ as follows: let $G_i$ be the mixed-radix topology induced by $\mathcal{N}_i$. Identifying the output nodes of $G_i$ with the input nodes of $G_{i+1}$ creates an $\overline{M}$-layer FNNT with ordered set $\mathcal{W}=(\mathbf{W}_1,\ldots,\mathbf{W}_{\overline{M}})$ of adjacency submatrices of the form (\ref{circshift}). Similarly, $\mathcal{D}$ implicitly defines a unique dense DNN topology $H$ on an ordered collection $U_0,\ldots,U_{\overline{M}}$ of nodes satisfying $\lvert U_i\rvert=D_i$. The ordered set of adjacency matrices of $H$ is $\mathcal{W}^*=(\mathbf{W}_1^*,\ldots,\mathbf{W}_{\overline{M}}^*)$, where $\mathbf{W}_i^*$ is the $D_{i-1}\times D_i$ matrix of ones. We define $G$ as the unique FNNT defined by
\begin{equation} \label{krons}
\overline{\mathcal{W}}=(\mathbf{W}_1^*\otimes\mathbf{W}_1,\ldots,\mathbf{W}_{\overline{M}}^*\otimes\mathbf{W}_{\overline{M}})
\end{equation}
(see Mathematical Preliminaries).

\emph{Properties of RadiX-Net Topologies:}

Sparsity:
We define the density of an FNNT $G$ with independent components $\mathcal{U}=(U_0,\ldots,U_n)$ as the ratio of the number of edges in $G$ to the number of edges in the unique dense DNN defined by $\mathcal{U}$. Here, we give a formula for the density of the RadiX-Net topology $G$ induced by $\mathcal{N}^*=(\mathcal{N}_1,\ldots,\mathcal{N}_M)$ and $\mathcal{D}=(D_0,\ldots,D_{\overline{M}})$. Let $\overline{\mathcal{N}}^*$ and $\overline{N}_i$ be as follows:
\begin{align*} \label{overline_n}
\overline{\mathcal{N}}^*&=(N_1^1,\ldots,N_{L_1}^1,N_1^2,\ldots,N_{L_2}^2,\ldots,N_{L_M}^M) \\
&=(\overline{N}_1,\ldots,\overline{N}_{\overline{M}})
\end{align*}
The density of $G$ is given by
\begin{equation} \label{density}
\frac{\sum_{i=1}^{\overline{M}}\overline{N}_iD_{i-1}D_i}{N^\prime\sum_{i=1}^{\overline{M}}D_{i-1}D_i}.
\end{equation}
One can approximate the density of $G$ as follows; let $\mu$ be the arithmetic mean of $\overline{\mathcal{N}}^*$, and assume that the variance of $\mathcal{D}$ is small. Then the density of $G$ is approximately $\frac{\mu}{N^\prime}$.

Symmetry:
By construction, RadiX-Nets satisfy symmetry (see Mathematical Preliminaries). In addition to guaranteeing path-connectedness, symmetry also serves to preclude any training bias that could be inherent in the underlying sparse DNN topology.

Path-Connectedness:
For any input node $u$ and any output $v$, the number of paths from $u$ to $v$ is equal to
\begin{equation} \label{count}
(N^\prime)^{\overline{M}-2}\left(\prod_{N\in\mathcal{N}_{M}}N\right)\left(\prod_{i=1}^{\overline{M}-1}D_i\right).
\end{equation}

\section{Discussion}

This paper presents the RadiX-Net algorithm, which deterministically generates sparse DNN topologies that, as a whole, are much more diverse than X-Net topologies while preserving their desired characteristics. In a related effort, benchmarking RadiX-Net performance in comparison to X-Net, dense DNN, and other neural network implementations can be found in \cite{alford}. Furthermore, RadiX-Net is used in \cite{wang} to construct a neural net simulating the size and sparsity of the human brain.

\section*{Acknowledgment}
The authors wish to acknowledge the following individuals for their contributions and support: Simon Alford, Alan Edelman, Vijay Gadepally, Chris Hill, Hayden Jananthan, Lauren Milechin, Richard Wang, and the MIT SuperCloud team.

\bibliography{RadiX-Net}
\bibliographystyle{ieeetr}

\end{document}